\documentclass[11pt,a4paper]{article}
\usepackage[hyperref]{eacl2021}
\usepackage{times}
\usepackage{latexsym}
\usepackage{times}
\usepackage{url}
\usepackage{times}
\usepackage{amsmath}
\usepackage{graphics}

\usepackage{graphicx}
\usepackage{footnote}
\usepackage{multirow}
\usepackage{algpseudocode}%
\usepackage{todonotes}
\usepackage{comment}
\usepackage{array}
\usepackage{tabularx}
\usepackage{arydshln}
\usepackage[linesnumbered,ruled]{algorithm2e}

\usepackage{microtype}

\aclfinalcopy 

\def \aclpaperid{344}
\title{\textit {DRAG}: Director-Generator Language Modelling Framework for Non-Parallel Author Stylized Rewriting}

\author{Hrituraj Singh \\
  Adobe Research\\
  \texttt{hrisingh@adobe.com}\And
  Gaurav Verma\thanks{This was work was carried out when the author was at Adobe Research.} \\
  Georgia Tech \\
  \texttt{gverma@gatech.edu}\\\AND
  Aparna Garimella \\
  Adobe  Research \\
  \texttt{garimell@adobe.com}\\\And
  Balaji Vasan Srinivasan \\
  Adobe Research\\
  \texttt{balsrini@adobe.com}\\
  }

\date{}

\begin{document}
\maketitle
\begin{abstract}
Author stylized rewriting is the task of rewriting an input text in a particular author's style. Recent works in this area have leveraged Transformer-based language models in a denoising autoencoder setup to generate author stylized text without relying on a parallel corpus of data. However, these approaches are limited by the lack of explicit control of target attributes and being entirely data-driven. In this paper, we propose a Director-Generator framework to rewrite content in the target author's style, specifically focusing on certain target attributes. We show that our proposed framework works well even with a limited-sized target author corpus. Our experiments on corpora consisting of relatively small-sized text authored by three distinct authors show significant improvements upon existing works to rewrite input texts in target author's style. Our quantitative and qualitative analyses further show that our model has better meaning retention and results in more fluent generations.
\end{abstract}

\section{Introduction}
With recent advances in language modeling techniques that have resulted in powerful language models \cite{radford2019language,devlin2018bert,brown2020language} along with an increased interest in stylized content generation, \cite{hu2017toward,shen2017style,subramanian2018multiple,fu2018style,niu2018polite}, large language models have been successfully tuned to achieve text stylization \cite{lample2018multiple,ziegler2019fine,syed2020adapting, singh2020incorporating}. Apart from \textit{transferring} an input text to the target style, which has received recent interest from the community,  \textit{understanding} and \textit{measuring} style have been persistently explored over the last few decades \cite{kessler1997automatic,garera2009modeling,liu2012sentiment,verma2019lexical}. Lying at the intersection of style transfer enabled by advanced language models and a deep understanding of style as a nuanced combination of several linguistic concepts, problems like \textit{stylized generation} or \textit{stylized rewriting} have gained further traction. 
 A large body of work in style transfer focuses on binary aspects such as positive-negative sentiment \cite{li2018delete,ziegler2019fine}, formal-informal \cite{jain2019unsupervised}, and sometimes a mixture of these attributes \cite{subramanian2018multiple}. To fuel this interest in such binary stylization, some datasets comprising of text from the extreme ends of these spectrums have also emerged (e.g.,  positive-negative sentiments \cite{mathews2016senticap}, formal-informal \cite{rao2018dear}). As pointed by \newcite{syed2020adapting}, author stylized rewriting does not directly fit under any of these variants as the writing style of an author is an amalgamation of several such attributes and needs to be modeled in a fine-grained manner. 

Apart from the distinction along style dimensions, prior works can also be categorized as supervised (using parallel corpus \cite{jhamtani2017shakespearizing}) and unsupervised \cite{li2018delete,syed2020adapting,niu2018polite}. 
In supervised frameworks, parallel data is used to tune sequence-to-sequence models for stylized rewriting. However, annotating such parallel corpus is a tedious effort and therefore, there is an increased interest in unsupervised style transfer; i.e., when there is no direct supervision or parallel data available for training the models. 
In this work, we focus on such an unsupervised setting.

Existing approaches on unsupervised author stylized rewriting rely on implicitly learning the target stylistic attributes from data and do not allow finer control on generation \cite{syed2020adapting}. While this is a good starting point for author-stylized rewriting, it is desirable to further improve the rewriting model on certain aspects without compromising on other attributes that the model has already optimized. An example would be to retain the stylistic strengths while improving content retention, or vice versa. To this end, we propose \textbf{D}i\textbf{r}ecting \textbf{a} \textbf{G}enerator framework (DRAG). Our quantitive and qualitative experiments show the viability of the proposed approach. Experiments further indicate that the framework's setup allows it to operate efficiently in scarce data setting and improves the performance over the baseline models.
Our contributions can be summarized as - \textbf{(1)} We introduce a director-generator approach to \textit{rewrite} an input text in a target author's style. \textbf{(2)} We propose linguistic alignment scores -- both at the local and global level and extend these to design \textit{thresholds} for the generator and director. \textbf{(3)} We present experimental results on texts written by three authors from the Gutenberg corpus with very distinct writing styles, and show that our approach outperforms prior works across content retention and style alignment metrics. \textbf{(4)} We further identify and discuss shortcomings of our proposed approach, and present error analysis to aid future research in author stylized rewriting.

\section{Related Work}
With the rise of Transformer-based \cite{vaswani2017attention} \textbf{language models}, generative pretraining \cite{devlin2018bert,radford2019language,brown2020language} has advanced the field of NLP significantly. Fine-tuning such large language models on specific task has become very prevalent \cite{sun2019fine,lee2020biobert,lample2019cross,raffel2019exploring,liu2019linguistic}. Pretraining infuses the \textit{generic} language knowledge into the language model helping it understand the specific tasks with relatively much less supervision. In fact, recent approaches \cite{radford2019language,brown2020language} show that often, even such small supervisions are not required and a simple instruction can be used to solve specific tasks by utilizing the capabilities of such large language models trained on very large datasets. 

Pretraining of such models usually involves optimizing them on \textbf{Masked Language Modelling (MLM)} \cite{devlin2018bert}, \textbf{Causal Language Modelling (CLM)} \cite{radford2019language} or other similar \cite{clark2020electra} objectives.
While CLM is the task of autoregressively predicting the next word given the previous words/context, MLM is the task of recovering masked tokens from a given input. While these approaches mostly train only an encoder or a decoder framework, \newcite{lample2019cross} explored initializing encoder-decoder frameworks using the pre-trained encoders for cross-lingual translation. Such a technique with appropriate modification has been shown to be successful in incorporating stylistic aspects of the language as well \cite{conneau2019cross,syed2020adapting}. All these works utilize the task of minimizing the \textit{denoising auto-encoder loss} for inducing style in the language models in a reconstruction framework. 
For our explorations, we leverage these works to initialize our {\sc DRAG} framework.  

There is an increased interest in \textbf{stylistic generation} or rewriting of content. Most of the approaches define dimensions like formality-informality \cite{shen2017style,ficler2017controlling,jain2019unsupervised,sun2019fine} and achieve the alignment along these dimensions. While some of these approaches rely on parallel corpus \cite{ficler2017controlling,jhamtani2017shakespearizing}, many of the approaches focus on unsupervised framework \cite{li2018delete,shen2017style,jain2019unsupervised}, where the model \textit{preserves }the input content in the output while \textit{biasing} the generations towards the target style. While some approaches utilize simple \textit{editing} to achieve the style along particular dimensions \cite{li2018delete}, others focus on achieving this through discriminators \cite{fu2018style} or scorers \cite{jain2019unsupervised}. As mentioned before, since author style is an amalgamation of several such attributes, it requires much more than a discriminator or singular dimension tuning to achieve stylization. 

Due to the difficulty associated with author style understanding and fine-grained nature of that style even if understood them, the problem of \textbf{author stylized rewriting} has not been explored a lot. While \newcite{jhamtani2017shakespearizing} try to solve this problem for a specific author (i.e. Shakespeare), their approach is contingent on the availability of a parallel corpus. Since preparing parallel corpus is a tedious and intractable process, especially while dealing with multiple authors and multiple combinations of input and output styles, it is essential to focus on unsupervised solutions. Most recently, \newcite{syed2020adapting} leverage the capabilities of the large language models to solve this problem in an unsupervised manner.

\section{Author Style}
There has been significant work on understanding binary stylization along dimensions like formal-informal, positive-negative sentiment \cite{rao2018dear,kessler1997automatic,pavlick2016empirical,collins2005predicting,hovy1990pragmatics,inkpen2006building,kantrowitz2003method}, however, there is limited work on understanding an author's writing style \cite{mccarthy2006analyzing,forgeard2008linguistic,verma2019lexical}.
While style can be a mixture of several factors including, but not limited to, lexical preferences, syntactic/sentential choices, discourse structure, narrative style, tone, we follow \newcite{syed2020adapting} and consider  an author's style at three levels:

\textbf{Surface style}  is estimated using the frequencies of different surface elements such as the number of commas, semicolons, colons, question marks, exclamation marks, and hyphens per paragraph, from a given author's text. We, thus, quantify the surface-style elements into a 6-dimensional vector.

\textbf{Lexical style} of an author is reflected in the author's choice of words. To describe the same concept, different authors may use different words. For instance,
Rudyard Kipling, known for his classics in children’s literature, tended to use more concrete words (e.g., gongs, rockets, torch) while Abraham Lincoln, being a political writer, used more abstract words (e.g., freedom, patriotism).
We enumerate lexical style categories as {subjective, objective, literary, colloquial, abstract} and  concrete \cite{brooke2013multi}.
We use lexicons for each of these categories \cite{brooke2013multi}, and define {\it lexical style alignment} of each word in the vocabulary to a given style category as the average and normalized point-wise mutual information (PMI) between that word and the seed words in the lexicon for that style category. The lexical style alignment for each word is thus a 6-dimensional vector. We use the EmoBank corpus \cite{emobank} to compute the co-occurrence statistics for PMI computations.
The inclination of a word towards a style category is positive if its normalized PMI score is positive with respect to the given category. The inclination of an author towards a style category is then estimated by the fraction of words in their text that have a positive inclination towards the category. 

\textbf{Syntactic style} of an author is indicated by the nature of sentences used and we estimate the distribution of different types of sentences in an author's text. Sentence types may range from complex, as seen in philosophical writings, to simple, as observed in children's storybooks.
We use five categories of sentence styles: {(i) simple, (ii) compound, (iii) complex, (iv) complex-compound sentences, and (v) others} \cite{feng2012characterizing,verma2019lexical,syed2020adapting}.
Sentences are categorized into one of these types using the algorithm proposed by \cite{feng2012characterizing}.
The resulting $5$-dimensional probability distribution vector is used as the estimation of syntactic style. These vectors are estimated at corpus-level, unlike those for lexical and surface style which are computed at paragraph-level.

\section{DRAG: \textit{D}i\textit{r}ecting \textit{a} \textit{G}enerator for Stylized Rewriting}

Our proposed framework, {\sc DRAG}, that aims to rewrite a given piece of text with a specific target author's style consists of three main stages:

\noindent{\bf (1)} Pretraining a language model to infuse general linguistic knowledge into the model

\noindent{\bf (2)} Adapting the pre-trained language model towards the target author's writing style by further pretraining it on text written by this author \cite{syed2020adapting}, and 

\noindent{\bf (3)} Using a director-generator framework (as discussed later) to fine-tune such \textit{biased} language model to improve its style transfer capabilities even further while fixing content preservation issues.
It is worth noting that we do not rely on the availability of parallel data for any of our experiments.
\subsection{Pretraining Language Model}
\begin{figure}[t]
    \centering
    \includegraphics[width=1.0\linewidth]{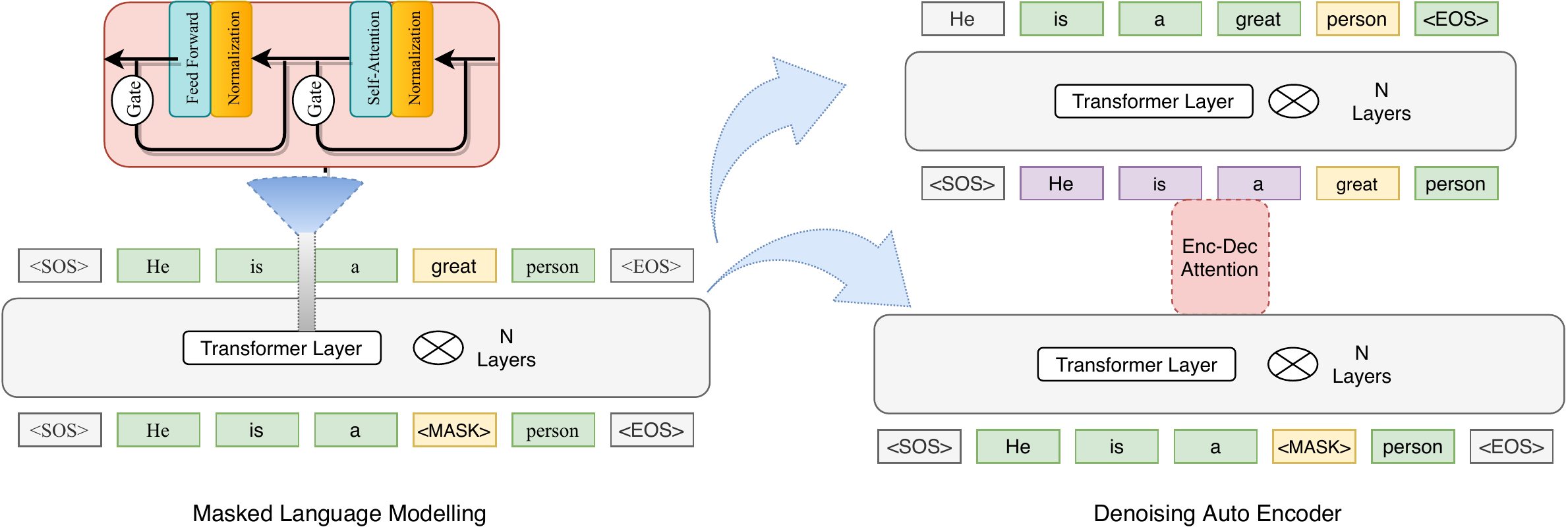}
    \caption{Language Model Pretraining using Masked Language Modelling followed by encoder-decoder initialization using pretrained models. This process still leaves the encoder-decoder attention parameters unitialized which can be initiliazed using the Denoising Auto Encoder training as depicted in the figure.}
    \label{fig:pretraining}
\end{figure}

In order to infuse general linguistic knowledge into a language model, we leverage Tranformer-based pretrained language models \cite{devlin2018bert,radford2019language,brown2020language}
due to their recent success in text processing tasks \cite{vaswani2017attention,devlin2018bert,brown2020language}.
Similar to \newcite{conneau2019cross}, we first train a Transformer-based encoder on the Masked Language Modelling (MLM) task with 15\% of the tokens masked \cite{devlin2018bert} on a {\it generic} text corpus. We initialize an encoder-decoder framework, as shown in Figure \ref{fig:pretraining}, with this language model. 


\subsection{Adapting LM for Rewriting}
{To adapt the pretrained LM for author stylized rewriting, \newcite{syed2020adapting} initialize an encoder-decoder framework with the pretrained LM, as shown in Figure \ref{fig:pretraining}. This is followed by optimizing it on denoising auto-encoder (DAE) loss \cite{lample2018multiple,lample2019cross} \textit{only} over target author's corpus. \newcite{syed2020adapting} use the DAE loss to infuse an author's linguistic style into the reconstruction model; we refer to this framework as \textbf{{\sc StyleLM}}. 
The fine-tuning using the DAE loss on a target author's corpus encourages recovering actual paragraphs from their noisy version \cite{lample2019cross}.} For a paragraph $g$ in corpus $G$ and its noisy version $C(g)$ ($C(.)$ being the noise function), DAE loss is given by,
\begin{equation}
\small
    DAE(\theta_e, \theta_{ed}, \theta_d) = -\frac{1}{|G|}*\sum_{g \sim G}\log{P(g/C(g);\theta_e, \theta_{ed}, \theta_d})
    \label{eq:dae}
\end{equation}
where $P$ is the probability of reconstruction for a given encoder parameters $\theta_e$, decoder parameters $\theta_d$, and encoder-decoder attention parameters  $\theta_{ed}$. Please note that $\theta_{ed}$ does not refer to any additional layer but the parameters which are present in transformers and are responsible for encoder-decoder attention. In our setup, $C(.)$ function introduces two noises: (a) random dropping of words with 10\% probability, and (b) word masking by replacing it with [MASK] token with 10\% probability.

Given a noisy input, the encoder fills the [MASK] tokens with suitable replacements (based on the knowledge from its MLM pretraining), thus creating a pseudo {\it generic} input for the decoder, the target sequence for which is aligned to the target author's style. However, we identify and verify experimentally two issues with this approach: 

\textbf{(1)} It requires a large target author corpora to achieve meaningful content preservation capability. This is evident by its very low content preservation scores (as discussed in Section \ref{sec:results}) when trained on authors with relatively smaller corpora. Even with large corpora, the model still suffers from exposure bias to texts written only by the target author leading to spurious outputs for \textit{unseen} inputs.

\textbf{(2)} The masking results in a significant emphasis on lexical style aspects, with a lesser focus on the surface and syntactic preferences. Since the model is completely data-driven, there is no way to explicitly add emphasis on additional style aspects. 

One of the primary reasons behind (1) is the lack of explicit initialization of encoder-decoder attention parameters in \textsc{StyleLM} resulting in a random initialization. The model, therefore, needs a large corpus of author data to stabilize these parameters. 
To fix this, we propose to train the entire encoder-decoder language model using the DAE loss over the same \textit{generic} corpus used for pre-training. 
The resulting model will be in the \textit{generic} language space (English, in our case), and henceforth referred to as \textsc{VanillaLM}.  We, further, finetune {\sc VanillaLM} in the author corpus on the DAE loss to arrive at an improved version of \textsc{StyleLM} which we call \textsc{iStyleLM}. This offers better encoder-decoder attention initialization, and also removes the exposure bias of {\sc StyleLM}, thus resulting in a more resilient and stable model with improved content preservation abilities (as demonstrated in Section \ref{sec:results}).

However, at this point,  we note that {\sc iStyleLM} still fails to address (2), and its content preservation ability is also sub-optimal as the target author's style aspects which are infused at the later stage of training override some of the general linguistic knowledge.
To further improve on {\sc iStyleLM}, we introduce a Director-Generator component to our training framework in the next section.

\begin{figure*}[!hbt]
    \small
    \centering
    \includegraphics[width=0.8\linewidth]{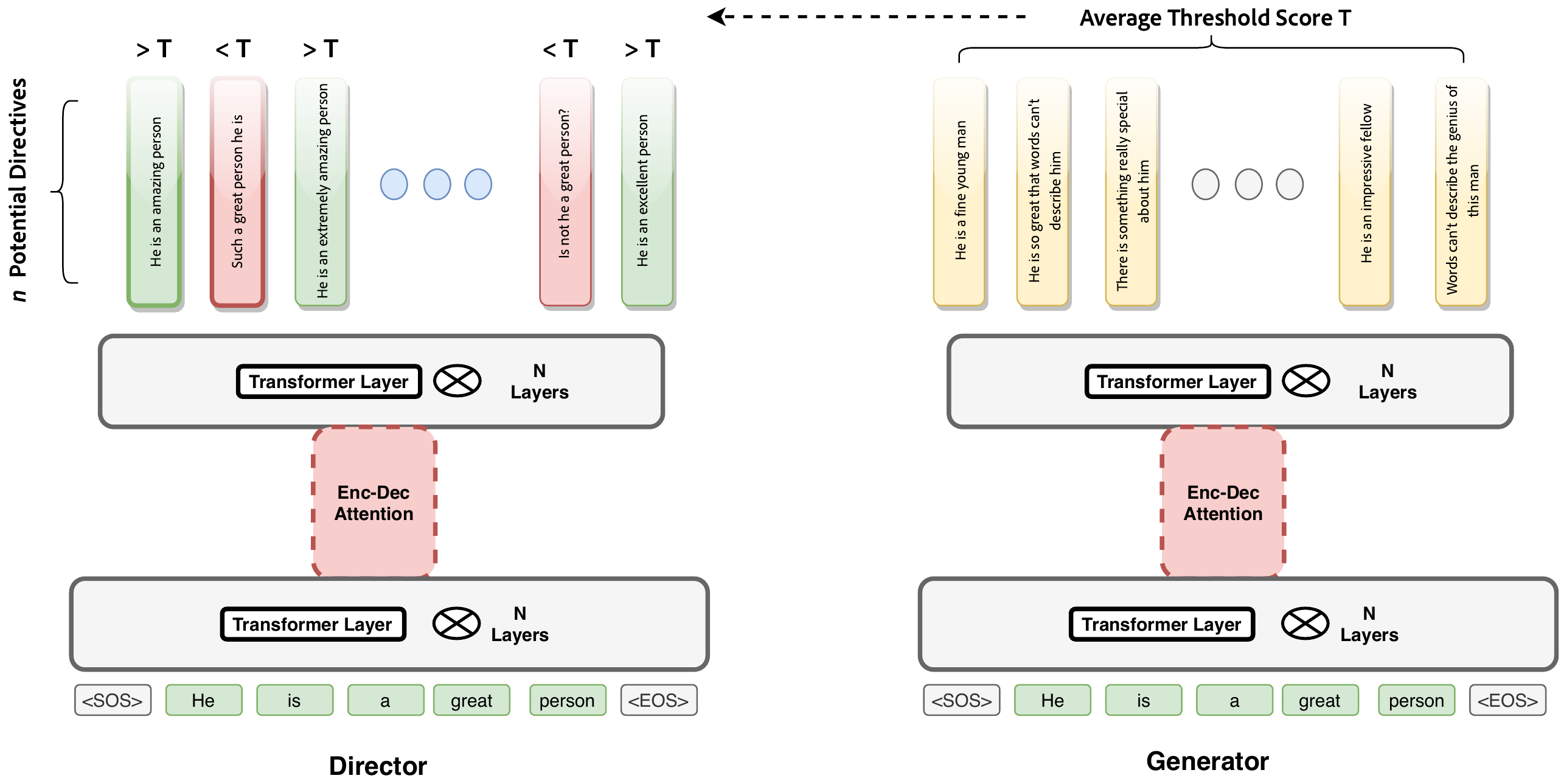}
    \caption{Both the director as well as generator, intiliazed using \textsc{iStyleLM}, work together to improve the final outputs. While director remains in the space of author style generating and exploring potential directives , generator keeps changing its threshold as it gets improved on its content \& style capabilities. The directives above the average threshold for same example are accepted while rest of them are rejected.}
    \label{fig:proposal}
\end{figure*}
\subsection{Director-Generator Finetuning} 
For the Director-Generator finetuning, we find inspiration in the standard RL strategies \cite{rennie2017self,ranzato2015sequence} where the \textit{nearby} space is \textit{explored} and certain actions are rewarded higher than others, consequently getting encouraged in the future. We, however, find direct rewarding unstable for our problem. Hence, we \textit{generate} potential \textit{directives} during exploration and \textit{accept} or \textit{reject} them on the basis of thresholds. A \textit{directive}, in our context, is as an output paragraph generated from an input by a director model which is fixed and has been initialized using \textsc{iStyleLM}.  Specifically, we create two copies of the \textsc{iStyleLM} as the {\it Director} and the {\it Generator}. As the names indicate, for each input, the director proposes $n$ potential directives or paragraphs, while the generator generates $n$ \textit{thresholding} outputs (paragraphs) as shown in Figure \ref{fig:proposal}. 
We generate the potential directives using nucleus sampling \cite{holtzman2019curious} with a softmax temperature of 1.2, while the thresholding outputs are generated using a softmax temperature of 0.8 (the same value is used at inference time as well). We score the director and generator outputs on various content and style attributes. For content preservation, we use the BLEU score between input and output as the content score. For lexical style, the mean squared error is calculated between the 6-dimensional lexical alignment vector of the directives/generator outputs (calculated as the averaged sum of alignments of words in the proposal) and average lexical alignments of paragraphs for the target author corpus. Similarly, the mean squared error for surface style is also calculated. The scores $L$ and $S$ for lexical and surface styles, respectively, are then calculated as reciprocal of means squared errors (with $\epsilon$ added in the denominator to avoid zero-division). For syntactical choices, since we wish to achieve the probability distribution of different types of sentences at the corpus-level, we calculate the score for syntactic style as, $SX = \frac{sum(P_p \circ P_t)}{sum(P_p)}$ where $P_p$ denotes the frequency distribution of different types of sentences in a directive/generator's output, $P_t$ the probability distribution of different types of sentences in target author corpus, and $\circ$ denotes the Hadamard product. All three scores are summed to calculate the style score for the directives (and the generator outputs).

The \textsc{iStyleLM} model already captures certain stylistic aspects of the target author. We want our model to leverage this understanding and improve on aspects where \textsc{iStyleLM} does not perform well. To capture this, we compute the content and style scores of all the potential directives and generator outputs and retain \textit{only} those directives which have \textbf{both} the content and style scores better than the average of the generators' outputs' scores.  
The \textit{accepted} directives become \textit{real} directives for the generator and are used to train it using the \textit{teacher-forcing} cross-entropy loss.
Note again that the director remains frozen with \textsc{iStyleLM} during the entire training process. In the case of multiple potential directives being better than the generators' outputs' average, the cross-entropy loss for each directive is weighted by its marginal difference from the generator's average score on the style dimension; i.e., if the style score for a directive is $D_s$ and average outputs' style score from the generator is $G_s$, its weight during the cross-entropy training is $D_s - G_s$. This objective is similar to the one used in SCST \cite{rennie2017self} but only \textit{accepted} directives are encouraged and nothing is explicitly discouraged. 

In order to stabilize the Director-Generator fine-tuning framework, we use (a) {\bf fixed director}, and (b) \textbf{moving generator}. Contrary to the natural expectation of exploring better directives with the training of the director, the fixed or frozen director prevents catastrophic degradation in case the training biases the model towards specific choices that further train the model. It is a known phenomenon in RL frameworks that the model quickly learns to bias towards specific choices that are more rewarding. Specifically, we observe that training the director as well leads to overfitting to the limited stylistic choices, thus resulting in the exploration of sub-optimal potential directives that seldom cross the required thresholds, especially the content preservation ones. With a moving (i.e. trained at each step) generator, its outputs scores account for the \textit{current} state of the model against a fixed stable director, and hence only those directives get accepted which are better than the \textit{current} capabilities (thresholds) of the generator. With a fixed generator, directives that would have been worse than \textit{current} capabilities of the model but better than the capabilities of the fixed generator would also get accepted, thus training the model in the opposite direction.  The Director-Generator finetuned \textsc{iStyleLM} yields our proposed {\sc DRAG} framework. At the inference time, we drop the director and use the Generator as our final rewriting model.

\section{Experiments}
\label{sec:experiments}
We use a transformer encoder\footnote{As proposed by \citeauthor{parisotto2019stabilizing} and shown in Fig. \ref{fig:pretraining}} with  512 hidden units, 16 heads, a dropout rate of 0.1, and
learned positional embeddings during our MLM training. The model is trained using Adam Optimizer with a learning rate of $10^{-4}$. The batch size used is 32 with a stream of 256 tokens, and the whole setup is trained until the validation performance (perplexity scores) shows no further improvement. The Transformers used in encoder-decoder setup also have the same parameters, and are initialized using the above encoder before training on further objectives. During DAE loss training, we use the same hyper-parameters used in \cite{conneau2019cross,syed2020adapting}, and set $p_{drop}$ and $p_{blank}$ to 0.1. During director-generator training, we use $n$ as 8 and $\epsilon$ as 0.05. The learning rate used in this case is $10^{-5}$. In all the models, we use Byte Pair Encoding \cite{sennrich2015neural} with 80$k$ codes learnt over the entire \textit{generic} corpus. 

\subsection{Dataset}
We use the 2,857 books written by 142 authors in the Gutenberg corpus \cite{lahiri}, as used in \cite{syed2020adapting}, along with the Wikipedia articles, to form a corpus of about 4.6M passages.
We refer to this corpus as \textit{generic} during all our experiments, since it infuses only \textit{generic} linguistic knowledge into the models.
While MLM and \textsc{VanillaLM} are trained on the \textit{generic} corpus, we select three authors with the most distinct writing styles, namely Albert Einstein, Michael Faraday, and John Stuart Mill, as measured by comparing their lexical alignments with the average lexical alignment of the Gutenberg corpus, as the target authors for author-specific style rewriting.
Note that the choice of the authors is made purely on statistical basis with these three authors having maximum lexical style difference on their style vectors as described earlier when compared with the lexical style of entire 
\textit{generic} corpus. 
For evaluation, we use the Opinosis corpus \cite{ganesan2010opinosis} as well as mixed author Gutenberg subset (with five passages from all the authors except the target author), which we refer to as \textit{Generic (Test)}.

\renewcommand{\arraystretch}{1.2}
\begin{table*}[bth]
    \centering
    \scalebox{0.68}{%
    \begin{tabular}{|c|c|c|c|c|c|c|c|c|}
    \hline
    \multirow{2}{*}{\textbf{Dataset}}&\multirow{2}{*}{\textbf{\textbf{Model}}}&\multicolumn{4}{c}{\textbf{Content Preservation} ($\uparrow$)}&\multicolumn{3}{|c|}{\textbf{Author Style} ($\downarrow$)}\\
    &&\textbf{ROUGE-1}&\textbf{ROUGE-2}&\textbf{ROUGE-L}&\textbf{BLEU}&\textbf{Lexical(RMSE})&\textbf{Surface(RMSE)}&\textbf{Syntactic(JSD)}\\
    \hline
    \multirow{4}{*}{\textbf{Opinosis}}& Vanilla LM & 75.23 & 56.12 &74.28 & 59.46 & 0.232 & 2.74 & 0.132\\ 
    &\textbf{StyleLM \cite{syed2020adapting}} & 32.61 & 14.51 & 31.69 &14.82 & \textbf{0.174} & 2.56 & 0.121 \\
    &\textbf{iStyleLM} & 49.28 & 29.16 & 49.86 & 28.16 & 0.178 & 2.61 & 0.122 \\
    &\textbf{DRAG} & \textbf{57.23} & \textbf{36.12} & \textbf{56.98} & \textbf{37.53} & 0.179 & \textbf{2.48} & \textbf{0.109 }\\
    
    \cdashline{0-8}
    
    \multirow{4}{*}{\textbf{Generic}}&Vanilla LM & 72.34 & 54.65 &71.93 & 56.46 & 0.218 & 2.48 & 0.120\\ 
    &\textbf{StyleLM \cite{syed2020adapting}} & 30.31 & 11.58 & 29.77 &12.91 & \textbf{0.163} & 2.29 & 0.114 \\
    &\textbf{iStyleLM} & 45.12 & 26.08 & 44.36 & 24.16 & 0.171 & 2.33 & 0.121 \\
    &\textbf{DRAG} & \textbf{52.39} & \textbf{30.66} & \textbf{51.98} & \textbf{33.28} & 0.174 & \textbf{2.21} & \textbf{0.097 }\\
    \hline
    \end{tabular}%
    }
    \caption{ $\uparrow$ indicates higher scores are better while $\downarrow$ indicates the opposite. Apart from lexical alignment where StyleLM performs marginally better, DRAG outperforms prior approaches. Vanilla LM performs best at content preservation but lacks any stylization}
    \label{tab:results}
\end{table*}

\begin{table*}[bth]
    \centering
    \scalebox{0.69}{
    \begin{tabular}{|p{5cm}|p{6cm}|p{5cm}|p{5cm}|}
    \hline
    \textbf{Input} & \textbf{Albert Einstein }& \textbf{Michael Faraday} & \textbf{John Stuart Mill}\\ 
    \hline
    The accuracy at this point is very good & The \textcolor{brown}{experimental definitions} developed is very \textit{clearly} & The point is very \textcolor{blue}{wonderful} & The physical for this \textcolor{red}{\textbf{,} }is \textcolor{red}{very very pretty} .\\
    \hline
    The estimated time to arrival does not seem to calculate the travelling time accurately & 
    The estimated time \textcolor{brown}{relative} to the leading \textcolor{brown}{existence} does not seem likely to calculate the travelling time \textit{exactly} & 
    The \textcolor{blue}{discovery} of \textcolor{blue}{ascertaining}  time \textcolor{blue}{\textbf{;}} \textcolor{blue}{indeed \textbf{,}} do not not show accuracy to the time to \textit{angles} & 
    The total time is to \textcolor{red}{infer} that arrival is not \textcolor{red}{verified} but often clearly a \textcolor{red}{\textbf{,}} accurately .\\
    \hline
    \end{tabular}}
    \caption{Qualitative Outputs For Three different authors for same inputs}
    \label{tab:q1}
\end{table*}
\begin{table*}[!bth]
    \centering
    \scalebox{0.75}{
    \begin{tabular}{|p{6cm}|p{7cm}|p{6.5cm}|}
    \hline
    \textbf{Input} & \textbf{StyleLM~\cite{syed2020adapting}}& \textbf{DRAG (Ours)}\\ 
    \hline
    but after that it is very \textbf{easy} and quite \textbf{accurate} to use. & But for all that it is very After question about this and quite \textcolor{red}{\textit{measured}} with consideration. & But after all it is very \textcolor{blue}{accurate} and quite \textcolor{blue}{illustrious} to the use of \textit{events}. \\
    \hline
    Leather seats are very comfortable. & come on very very should we have any replaced. & This \textcolor{blue}{moving \textit{hypothetical} seats} are very comfortable. \\
    \hline
    I am not real fond of the electric seat and I find it is not as comfortable as my F150 pickup on trips & I am not real and use of the electric \textcolor{red}{\textit{position}} and I find that it is not as well may's for the very hardly small have we led train & I am not real fond of the electric seat \textbf{\textcolor{blue}{,}} and I find it is not as comfortable as \textcolor{blue}{\textit{my physical relative on railway investigations. }}. \\
    \hline
    \end{tabular}}
    \caption{Comparison between \textsc{StyleLM} and \textsc{DRAG} for Albert Einstein}
    \label{tab:q2}
\end{table*}
\subsection{Quantitative Evaluation}
\label{sec:results}
Table \ref{tab:results} shows the results averaged over the three selected authors.
The experiments are conducted on Opinosis and {\it Generic (Test)} datasets, using the following four models.
\begin{itemize}
    \item \textbf{\textsc{VanillaLM}} is initialized using MLM-trained encoders and decoders and fine-tuned on the generic corpus using DAE loss. 
    \item \textbf{\textsc{StyleLM}}, proposed by \newcite{syed2020adapting},\footnote{Note that the {\sc StyleLM} code is not publicly available. Results shown in the table are using our own implementation.} is also initialized using MLM-trained encoders and decoders, but fine-tuned \textit{only} on the target author corpus (instead of the generic corpus). 
    \item \textbf{\textsc{iStyleLM}}, an improved and stronger baseline compared to {\sc StyleLM}, is initialized with \textsc{VanillaLM} and then fine-tuned on the target author corpus. 
    \item \textbf{\textsc{DRAG}} is our proposed model. We use {\sc iStyleLM} to initialize both director and generator as described above, and then fine-tune them using inputs from \textit{generic} corpus.
\end{itemize}
While the \textit{Generic(Test)} corpus is predominantly literary due to the nature of the source, Opinosis covers everyday language. As shown in Table \ref{tab:results}, \textsc{StyleLM} improves on the style alignment scores, but at a great cost of content preservation when the target author corpus is small. This is possibly due to the random initialization of encoder-decoder attention parameters in the DAE training over target corpus, as reflected in the superior performance of \textsc{iStyleLM}. 
We also note that while the approach proposed in \textsc{StyleLM} \cite{syed2020adapting} improves lexical scores significantly, it fails to bring the same level of improvement in surface and syntactic alignments, perhaps due to the due to rare chances of less frequent punctuation symbols getting masked during DAE training, even more so when the target author corpus is not large enough to cover all possible masks. Similar reasoning explains the syntactic alignment issues, The \textsc{DRAG} approach, however, improves on both surface and syntactic alignment along with content preservation scores even though it comes at the marginal cost of lexical alignment. Please note that the purpose of Vanilla LM is to provide an estimation of  \textit{upper limit }on the content preservation scores and is not to be treated as a baseline due to the simple objective of its task (just copying the input tokens).

\subsection{Qualitative Comparisons}
We also qualitatively show some comparisons for different authors and different models. In Table \ref{tab:q1}, we show the outputs of {\sc DRAG} for same input and different target authors. Evidently, our model produces changes both at the lexical as well as surface levels. Word 'good' in the first input is replaced by words like 'clearly', 'wonderful' and 'pretty' depending on the author. Some words do not replace any word but still get added to change the syntactical structure of the sentences. For example - appearance of word 'relative' starts comparison to the 'leading existence' making it a bit complex. Sometimes surface level changes like appearance of ';' also change the complexity of sentences.

We also show the comparison between \textsc{StyleLM} and our proposed \textsc{DRAG} outputs for same inputs when the target author is Albert Einstein as shown in Table \ref{tab:q2}. Evidently, while both models try to achieve the stylistic alignment, \textsc{StyleLM} ends up distorting the input sentence too much resulting in poor content preservation properties. Words like 'measured', 'hypothetical', and 'physical' relative reflect the \textit{objective} approach used in Albert Einstein's writings. 

\section{Discussions and Limitations}
While the language generation advancements are happening at a very high pace, the notion of style and the ability of models to rewrite same content in different styles is still far from being solved. One of the most important observation as made by \cite{lample2018multiple} is that it is very difficult to separate content from style. In fact, previous approaches which worked on the principle of \textit{disentangling} style from content were not found to disentangle the style so much after all \cite{lample2018multiple}. The notion of style is still very far from being defined and concretized. While some psycholinguistic concepts can be defined to some extent (formality, sentiment, etc.), defining it at the level of author's style is very difficult due to manifestation of style at different levels as enumerated by \cite{verma2019lexical}. Despite, such enumeration at various levels, it is far from \textit{exhaustive} and therefore our approach still requires more granular understanding of style to closely emulate target author's style.

Our evaluation uses automatic metrics for style due to the difficulty associated with conducting human evaluation in author attribution tasks\cite{syed2020adapting}. The skill needed to identify the author's style is very intense thus making the human evaluation very costly. A more granular and detailed study on understanding how humans interpret an author's style is required to design a proper feedback mechanism. This is, however, outside the scope of this work.

\subsection{What Did Not Work}\label{sec:failure}
In this section, we discuss some of our explorations that did not work as expected to aid future research in author stylization. We experimented with various \textbf{reinforcement learning setups} as it was a more natural choice once we had scoring engines for rewards. Using the \textsc{VanillaLM} as a policy and we explored Self Critical Sequence Training (SCST) \cite{rennie2017self,ranzato2015sequence} and Proximal Policy Optimization\cite{schulman2017proximal}. However, all the setups were unstable in various ways for our problem. Note that our experiments and observations here are limited to the problem of author stylized rewriting only. SCST or self-critical sequence training is aimed at bringing the advantages of reinforcement learning setups for sequence level problems. A model (or policy) generates/explores outputs (or episodes) using multinomial sampling and greedy sampling. If the greedily sampled episode reward is $r_b$ and the non-greedily sampled episode reward is $r$ - the whole setup is trained using REINFORCE\cite{sutton2018reinforcement} with $r$ as the actual reward and $r_b$ as baseline reward. We found this to limit exploration considering our problem is relatively much harder than previous metrics on which SCST has been successful due to our target metric being of an exact value. It, therefore, resulted in no improvement in either style or content scores. We, therefore, shifted to its modified version to encourage exploration, where we generated multiple episodes for each input and averaged their scores to use that as baseline reward $r_b$ and trained the setup on all generated episodes using REINFORCE \cite{sutton2018reinforcement}. We found this approach to be effective at style incorporation but not generalizable at all. The model learned to repeat certain patterns with poor content preservation abilities. We, tried, to balance it with occasional denoising autoencoder loss training but that only \textit{delayed} the overfitting and not solve it. We also attempted Proximal Policy Optimization in a setup same as \cite{sun2019fine} but it resulted in even worse outputs due to the critic's failure to approximate complex value functions for our objectives.

As discussed already, we only accept those directives which have scores above the threshold. We also tried a variant of it which had even those directives which do not score above the threshold. We scored them negatively thereby resulting in a bit similar framework like SCST but within some steps, we found more negative scores than positive due to bad content preservation pushing the model away from a bad state towards some undefined state resulting in spurious and inconsistent outputs.

\section{Conclusion and Future Work}
In this work, we addressed the shortcomings of the prior approaches for the task of author stylized rewriting and overcame them through {\sc DRAG}: a Director-Generator approach. We showed the effectiveness of our proposed approach for stylized rewriting on three different authors from the Guteneberg Corpus. Furthermore, we discussed the limitations of our approach and some of the failure cases to aid future research. While our DRAG approach is able to stabilize the training while improving the content preservation abilities of the model, a standard reinforcement learning approach, when stabilized, has the potential to improve these scores to a much more improved level. Improved understanding of author style while keeping a human in the loop and stabilizing RL with transformers models are subjects of future research.
\bibliographystyle{acl_natbib}
\bibliography{eacl2021}
\end{document}